\pdfoutput=1

\documentclass[11pt]{article}

\usepackage[review]{acl}

\usepackage{times}
\usepackage{latexsym}
\usepackage{enumitem}
\usepackage{booktabs}
\usepackage{amssymb}
\usepackage{amsmath,graphicx,amsfonts}
\usepackage{graphics}
\newcommand{\newcheckmark}{\raisebox{0.6ex}{\scalebox{0.7}{$\sqrt{}$}}}
\newcommand{\newcrossmark}{\scalebox{0.85}[1]{$\times$}}

\usepackage[T1]{fontenc}

\usepackage[utf8]{inputenc}

\usepackage{microtype}

\title{Speech-based Slot Filling using Large Language Models}

\author{Guangzhi Sun$^{1}$, Shutong Feng$^{2}$, Dongcheng Jiang$^{1}$, Chao Zhang$^{3}$,\\
\bf{Milica Ga\v{s}i\'{c}}$^{2}$, \bf{Philip C. Woodland}$^{1}$ \\
  $^1$University of Cambridge Department of Engineering \\
   $^2$Heinrich Heine University D\"{u}sseldorf \\
   $^3$Tsinghua University \\
  \texttt{gs534@cam.ac.uk} \\}

\begin{document}
{\makeatletter\acl@finalcopytrue
  \maketitle
}
\begin{abstract}
Recently, advancements in large language models (LLMs) have shown an unprecedented ability across various language tasks. This paper investigates the potential application of LLMs to slot filling with noisy ASR transcriptions, via both in-context learning and task-specific fine-tuning. Dedicated prompt designs and fine-tuning approaches are proposed to improve the robustness of LLMs for slot filling with noisy ASR transcriptions. Moreover, a linearised knowledge injection (LKI) scheme is also proposed to integrate dynamic external knowledge into LLMs. Experiments were performed on SLURP to quantify the performance of LLMs, including GPT-3.5-turbo, GPT-4, LLaMA-13B and Vicuna-13B (v1.1 and v1.5) with different ASR error rates. The use of the proposed fine-tuning together with the LKI scheme for LLaMA-13B achieved an 8.3\% absolute SLU-F1 improvement compared to the strong Flan-T5-base baseline system on a limited data setup.
\end{abstract}

\section{Introduction}
\label{sec:intro}
Slot filling, as an important sub-task of spoken language understanding (SLU), is a crucial component in conversational AI such as spoken dialogue systems. It requires the extraction and understanding of pertinent information in the user's speech query. Accurate extraction of slot values from the query speech is indispensable for accurate response generation and is challenging with limited annotated data and noisy ASR transcriptions. In particular, domain-specific named entities that are crucial to accurate information extraction, usually have high error rates with a generic ASR system. Although this problem can be mitigated by training systems on data in the target domain, it can be expensive to construct dedicated large-scale training data for a specific SLU task as it requires extensive labelling effort and domain expertise \cite{fewshotsf1,fewshotsf2,BERT4SFlowdata}. This data sparsity problem can be addressed by transfer learning with pre-trained language models (PLMs) \cite{QASF,QASFzero}, especially with large language models (LLM).

Recent advancements in LLMs, such as GPT-4 \cite{instructgpt,gpt4} and the LLaMA series \cite{llama,llama2}, have been shown to exhibit human-level reasoning ability for natural language tasks even without task-specific fine-tuning of the model parameters, which is known as the \textit{emergence} of LLMs. This is usually achieved by conditioning the model generation on a prompt containing examples or a task description, referred to as in-context learning. Although demonstrably effective, studies have also shown that LLMs struggle with accurate fine-grained content extraction such as slot-filling \cite{llmdialogue,llmdialogue2,llmsf,llmsf2}, and tend to overly extrapolate beyond the examples and task descriptions in the prompt, especially when evaluated using text-based quantitative metrics. This necessitates the use of dynamic contextual knowledge to guide and confine the generation \cite{llmknowledge1,llmknowledge2}, as well as efficient task-specific fine-tuning with limited data \cite{llmfinetune1}. Moreover, as slot-filling heavily relies on the quality of ASR \cite{cti,rnntinterface,aedinterface1,TCPGenSLU}, it is essential to improve the performance of LLMs with noisy ASR transcriptions. 

This paper aims to quantify the performance of LLMs for slot filling with different ASR error rates and proposes prompt designs and effective fine-tuning methods with the aid of external dynamic knowledge to perform SLU using noisy ASR transcriptions as the input. Specifically, the GPT-3.5-turbo and GPT-4 models are selected as state-of-the-art LLMs for in-context learning evaluation, while LLaMA and Vicuna are adopted as two widely-used open-source LLMs at a smaller scale that can be finetuned locally on task-specific data. LLaMA is the basic LLM without any dedicated fine-tuning, whereas Vicuna is obtained after fine-tuning on user-shared conversations with GPT-3.5. On the ASR side, this paper focuses on generic open-source large speech models and uses a group of off-the-shelf Whisper models \cite{whisper} to provide noisy transcriptions at different word error rate (WER) levels. In addition to the one-best hypothesis, the proposed prompt design is also able to incorporate multiple hypotheses in the form of an $N$-best list for a single utterance to improve the model's robustness to ASR errors. Moreover, by leveraging a pre-defined knowledge base (KB) \cite{TCPGenSLU}, dynamic knowledge found in the $N$-best list is also linearised into text and used in the prompt to provide necessary constraint that guides the language generation\footnote{Code and prompt template will be available at [URL]}.

Experiments were performed for the slot-filling task using SLURP data, where particular attention was paid to the performance of in-context learning and few-shot learning. The experimental results showed that instruction-tuned LLMs outperformed the baseline GPT-2 model and Flan-T5-base model in both in-context learning and limited data fine-tuning setups. The main contributions of this paper can be summarised as follows.
\vspace{-0.3cm}
\begin{itemize}[leftmargin=10pt]
\setlength\itemsep{-0.5em}
    \item The performance on slot filling is determined using SLURP data for a range of widely used LLMs, including GPT-3.5-turbo, GPT-4, LLaMA and Vicuna.
    \item A prompt design and data-efficient noise-robust fine-tuning approach for slot filling using LLMs with noisy ASR transcriptions is provided.
    \item A linearised knowledge injection (LKI) scheme is proposed that incorporates contextual knowledge derived using $N$-best ASR hypotheses into the prompt for LLMs.
\end{itemize}


\section{Related Work}
\label{sec:relwork}

\subsection{LLMs}
LLMs refer to the type of LMs with billions of model parameters and trained on vast amounts of data. GPT-3 (175 billion parameters) \cite{gpt3} and PaLM (540 billion parameters) \cite{palm} were early examples of LLMs, significantly outperforming their predecessors such as BERT \cite{bert} (330 million parameters) and GPT-2 \cite{gpt2} (1.5 billion parameters). One of the most prominent applications of LLMs is ChatGPT, which was built from GPT-3.5 via reinforcement learning with human feedback (RLHF) \cite{NIPS2017-d5e2c0ad} and was adapted for chat applications. Later, a larger scale LLM, the GPT-4 \cite{gpt4}, was built to further support the visual modality. While the capability of LLMs continued to expand with ever-growing model sizes, ``smaller'' LLMs (still with tens of billions of parameters) such as LLaMA \cite{llama} were released and achieved a better balance between the performance and the computing resource required. Since LLaMA is open-source, many versions of LLaMA such as Vicuna \cite{vicuna} were developed with different conversation-based fine-tuning schemes. 

With scaled-up model and training data sizes, LLMs have demonstrated superior abilities in solving assorted complex tasks over their predecessors, which are known as ``emergent abilities'' \cite{emergent}. One of the key capabilities of LLMs is in-context learning, which has enabled LLMs to perform specific tasks by providing task descriptions or examples without explicitly updating model parameters. Early research explored in-context learning by providing the schema of the ontology in a spoken dialogue system. Specifically, in \cite{schema}, domain-slot relations from the dialogue ontology were encoded using a graph neural network (GNN) to guide the system. Later, \cite{hu-etal-2022-context} proposed an in-context learning framework by summarising the dialogue history into a short description, together with the schema description, as the context. More recently, \cite{instructgpt} proposed a more powerful LLM trained via RLHF that performed response generation via in-context learning.

\subsection{Slot Filling with Limited Data}

As slot-filling often requires domain expertise for labelling which makes it very expensive, data efficiency has been a crucial research topic. Fine-tuning PLMs on a relatively small-scale task-specific dataset for slot-filling has been one of the main themes in this area. In particular, compared to sequence tagging using pre-trained bi-directional encoder systems\cite{BERT4SFlowdata}, by formulating slot-filling as a sequence generation task, the power of language generation of LMs, such as GPT-2 \cite{QASFGPT2} or T5 \cite{T5,QASFT5, MultiwozZero2}, can be further exploited. These models achieved few-shot learning by only presenting a limited number of data samples, and it was also discovered that providing task descriptions in the context helps few-shot learning. Meanwhile, integrating prior knowledge about slots, such as possible values of each slot, also showed improvements in performance \cite{knowledgedial1, knowledgedial2}, especially with limited data\cite{TCPGenSLU}.

LLMs have enabled unprecedented slot-filling performance without task-specific fine-tuning by only presenting the LLM with a task description and several examples. The LLM will generate the desired slot and value pairs in the standard LM fashion \cite{llmsf,llmsf2,llmdialogue}. Structured contextual knowledge can be linearised into plain text as a part of the context to further improve in-context learning ability \cite{linknow}. Despite this ability and the performance achieved, the quantitatively measured slot-filling performance across a wide range of domains is still far from that of state-of-the-art fine-tuned systems \cite{llmsf, llmsf2}. 

\section{Methodology}
\label{sec:method}

\begin{table*}[t]
\caption{Prompt design for slot filling using LLMs, including task definition, in-context examples, linearised knowledge injection (LKI) scheme, query and the extra constraint for LKI only used for in-context learning. The example shows $K=1$ in this table.}
    \centering
    \footnotesize
    \begin{tabular}{lp{10cm}}
    \toprule
    Prompt parts     &  Prompt template \\
    \midrule
    Task definition & Consider the following list of slot types provided to you in JSON format:\\
    & \{``\texttt{slot A}": ``definition of slot A", ``\texttt{slot B}": ``definition of slot B", ...\}\\
    \midrule
    In-context examples (optional) &  For example, given ``\texttt{utterance\_1}" you should extract \{``\texttt{slot A}": ``\texttt{value A}"\}, \\
    &given ``\texttt{utterance\_2}" you should extract \{``\texttt{slot B}": ``\texttt{value B}"\}, etc.\\
    \midrule
    LKI (optional) & First, possible values for some slots are provided in the following KB: \{``\texttt{slot A}": ``\texttt{value A}", ...\}\\
    & Slot values not in the KB are not likely to be the correct value\\
    \midrule
    Query & Now consider the following sentence(s) containing one or more of the above slot types. \\ 
    & Extract slots belonging to that list and their values in JSON format.\\
    & i.e. \{``slot type": ``slot value"\}, or \{\} if no slots found \\
    \midrule
    Additional query with LKI & If, for a slot in KB, you can not find a value in KB, select 1 or 2 most likely values from KB instead \\
    \bottomrule
    \end{tabular}
    \label{tab:prompt}
\end{table*}
\subsection{Prompt Design for Slot Filling}
\label{sec:prompt}

The prompt design is shown in Table \ref{tab:prompt}, which comprises five major parts: \textit{task definition}, \textit{in-context examples}, linearised knowledge injection (\textit{LKI}), \textit{query} and \textit{additional query with LKI}. 

The task definition gives a brief definition for each slot type. It is necessary to encourage the generated response to focus on the slot types that are needed. The in-context examples provide further context to improve the performance. This part provides few-shot examples which comprise pairs of utterances and desired outputs in the prompt. Although in-context examples provide effective guidance on what to generate, the number of samples in the prompt is always limited by the context length. The LKI and the additional query with LKI are two parts that appear in pairs when an external KB is used. As LLMs tend to extrapolate and create an excessive number of slot-value pairs, these two KB-related parts are used to provide a strong prior that confines the generation of certain slots to the allowed set of possible values. The main query prompts LLMs with the question, followed by utterances from which slot values are extracted. The basic form of the prompt is the concatenation of the task definition and the query, followed by the utterance, with the other three parts being optional.

\begin{figure*}[t]
    \centering
    \includegraphics[scale=0.26]{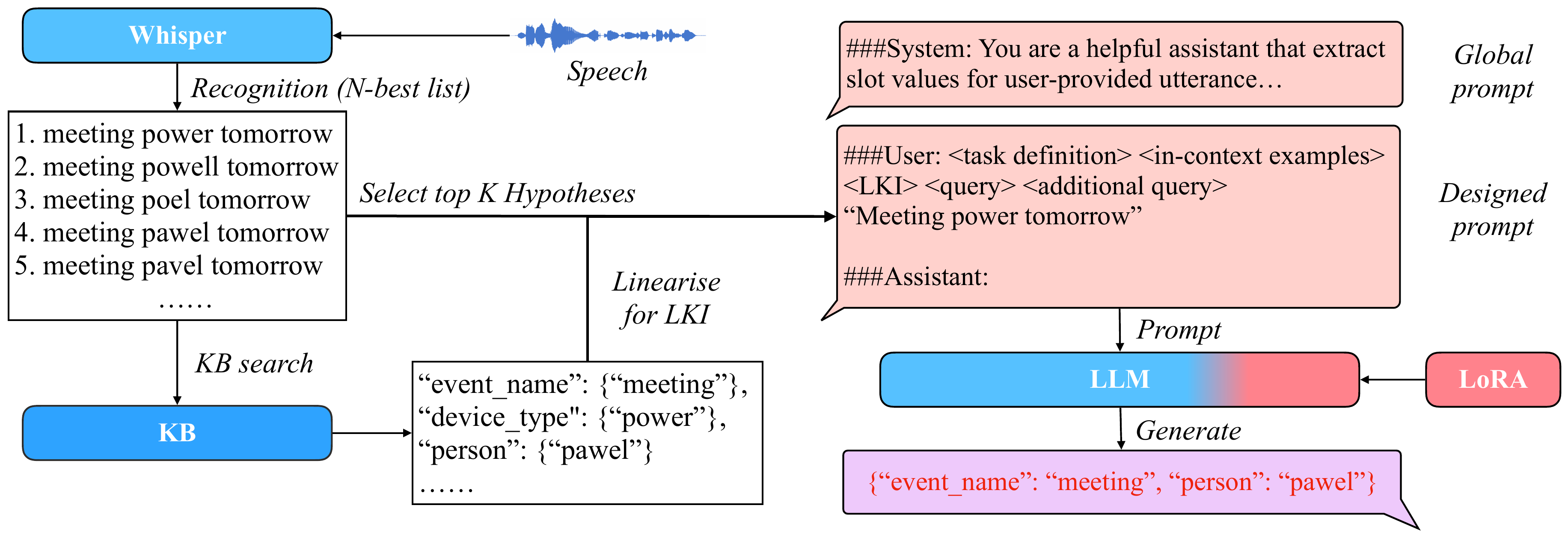}
    \caption{Diagram illustrating the slot-filling pipeline using Whisper and LLM. The user prompt comprises parts introduced in Table \ref{tab:prompt}, followed by the top one hypothesis in the $N$-best list as an example. Low-rank adaptation (LoRA) is used for fine-tuning open-source LLMs. Note that the specific format of the roles in the prompt, e.g. \#\#\#Assistant, is dependent on the specific LLM.}
    \vspace{-0.3cm}
    \label{fig:flowchart}
\end{figure*}

The complete pipeline for slot filling with LLMs taking speech as input is shown in Fig. \ref{fig:flowchart}. Starting with the speech input, a Whisper model is used to transcribe the speech into an $N$-best list. The prompt is then constructed as designed, with an additional global prompt for systems with instruction tuning. The top $K$ hypotheses, where $K=1$ by default, are used as the final part of the prompt. The complete prompt is sent to the LLM to generate a text sequence that completes the assistant response using search algorithms as shown in Eqn \eqref{eq:search}.
\begin{equation}
    \mathbf{W}^* = \underset{\mathbf{W}}{\arg\max} P(\mathbf{W}|\mathbf{W}^P),
    \label{eq:search}
\end{equation}
where $\mathbf{W}$ is the generated token sequence and $\mathbf{W}^P$ is the prompt token sequence. Note that there is no stochasticity in the generation process as the task is to extract exact information from an utterance for quantitative measurements. 

The LKI part of the prompt is proposed to improve the robustness of LLMs by constraining the generation with a pre-defined KB, especially with noisy transcriptions. The external KB contains all possible named entities for each slot type. Then, for each utterance, entities that can be found in the $N$-best list via string matching are selected together with their slot types, as shown in the example in Fig. \ref{fig:flowchart}. The selected slot values are linearised into a text format to be used as part of the prompt. The use of LKI not only provides guidance in the generation process to generate sensible values constrained by the KB but also makes further use of ASR alternatives. For example, the name ``pawel" in the utterance has different substitutions (e.g. ``power"), and when using the top hypothesis for slot filling, the LLM is unable to fill this slot with the correct value. However, the name is correctly recognised in the 4th-best hypothesis and provided the entry in LKI. With that extra information in the prompt, the LLM has a much better chance to correctly perform slot filling for that entity.

\subsection{Task-specific Finetuning}
\label{sec:finetune}

Task-specific finetuning is particularly useful when a small amount of training data is available to leverage the few-shot learning ability of LLMs. The low-rank adaptation (LoRA) \cite{lora} finetuning method is used in this paper.

LoRA approximates the necessary update to a full-rank parameter matrix using a low-rank matrix, which can be decomposed into the multiplication of two low-rank matrices. Specifically, for a pre-trained matrix $\mathbf{W}_0\in \mathbb{R}^{m\times n}$ from a specific attention block, its update during finetuning can be constrained in the form shown in Eqn. \eqref{eq:lora}.
\vspace{-0.1cm}
\begin{equation}
    \mathbf{W}_0 + \Delta \mathbf{W}= \mathbf{W}_0 + \mathbf{A}\mathbf{B}
    \vspace{-0.1cm}
    \label{eq:lora}
\end{equation}
where $\mathbf{A}\in \mathbb{R}^{m\times r}$ and $\mathbf{B}\in \mathbb{R}^{r\times n}$ contains trainable parameters with the pre-trained parameters, $\mathbf{W}_0$, fixed, and $r \leq \min (m, n)$. When setting $r$ to be a value much smaller than the model dimension, e.g. 8, the total number of parameters to be learned during fine-tuning is much smaller, which both improves the training efficiency and avoids over-fitting with a large number of parameters, especially for LLMs. Following \cite{lora}, LoRA is applied to the projection matrices of the self-attention layer in this paper.

LLMs are trained in the standard LM fashion with the cross-entropy (CE) loss applied to the generated output, i.e. the slot-value pairs in JSON format in Fig. \ref{fig:flowchart}. As the ASR system is an off-the-shelf Whisper model which is not fine-tuned in this paper, and, given that the training and validation set WER are similar, the $N$-best list derived from the training audio can be used to improve the robustness of LLM to the noisy transcriptions. Specifically, instead of feeding in the reference transcription, the top $K$ most likely hypotheses can instead be concatenated into one single string and used in the prompt during fine-tuning so that the model learns to robustly extract information from multiple hypotheses. In addition, LKI can be used during fine-tuning so that the model is aware of the knowledge provided and learns to use it. Random distracting entities can also be added to the LKI part to avoid the model being over-confident about the accuracy of the provided knowledge.

\section{Experimental Setup}
\label{sec:setup}
\subsection{Data}
{SLURP} \cite{slurp} is a collection of 25k single-turn user interactions with a home assistant, annotated with scenarios, actions and entities. Experiments were performed in two data setups. First, the official training, validation and test split following \cite{TCPGenSLU} was used where synthesised audio files were also included. Note that subsets of the training set were selected, with e.g. 2000 samples corresponding to 8\% of the full data, to demonstrate the limited data scenarios. 
In addition, a simulated zero-shot setup \cite{TCPGenSLU} was used. In training, all utterances containing entities of five randomly selected `unseen' slots were held out, and the held-out set was used for testing. 

The KB was organised as a simple dictionary for SLURP, where keys were slot types and values were lists of possible named entities for that type. It was created by collecting named entities that appeared in the entire SLURP data for each slot type, including train, validation and test sets, as a simulation of a real-world task environment. The average size of these lists was 106: the largest list was \texttt{person} which contained 872 entities, and the smallest list was \texttt{transport\_agency}, which only contained 2 entities.

\subsection{Models}
Five different LLMs were selected for evaluation in this paper: GPT-3.5, GPT-4, LLaMA-13B, and Vicuna-13B (v1.1 and v1.5). The first two models were chosen as the two most popular LLMs representing state-of-the-art performance of in-context learning without parameter fine-tuning. The other models were chosen as widely used open-source models that allow task-specific fine-tuning. Each model is introduced as follows:

\textbf{GPT-3.5} was used in ChatGPT, a chatbot application introduced by OpenAI. Although exact details about its architecture and training procedures are not published, according to OpenAI, it used a similar architecture to \citet{instructgpt} (175B parameters) and was trained with RLHF. The version released on March 1st, 2023 (GPT-3.5-turbo-0301) was used for the experiments.

\textbf{GPT-4} went one step further from GPT-3.5 by having a larger scale (1050B parameters) and additionally accepting images as input, though only the text modality was explored in this paper. In the experiments, the version released on March 16th, 2023 (GPT-4-0316) with a maximum sequence length of 4096 was used.

\textbf{LLaMA-13B} is an LLM with a Transformer decoder structure containing 13B parameters and was pre-trained on one trillion tokens. It took advantage of various techniques used in training other LLMs such as pre-normalisation \cite{prenorm}, the SwiGLU activation function \cite{swiglu}, and rotary embeddings \cite{rotaryemb}.

\textbf{Vicuna-13B} (version 1.1) is a fine-tuned version of LLaMA-13B which is adapted specifically to chat applications. It was trained on user-shared conversations with ChatGPT collected using a tool called ShareGPT. It also has 13B parameters. The upgraded version, \textbf{Vicuna-13B-v1.5}, derived from the LLaMA2-13B model was also used in this paper. The version supporting 16k tokens was used to represent the long context model.

As baselines and performance references, the GPT-2 model with 117 million parameters and the FLAN-T5 base model with 250 million parameters were employed that were fine-tuned on SLURP with all parameters.
Four different sizes of English Whisper models including tiny (39M), base (74M), small (244M) and medium (769M), were used as the off-the-shelf ASR models. The Whisper large model achieved very similar performance to the medium model and for efficiency, the medium model was used in the experiments. Both the reference and ASR transcriptions were normalised following the text normalisation scheme in \cite{whisper}, and were scored against a normalised SLURP label file.

\begin{table*}[t]
\footnotesize
\caption{SLU-F1 (Precision/Recall) of LLMs with transcriptions from different Whisper ASR systems on SLURP test set. WERs of Whisper systems are also provided. LLMs were evaluated using task descriptions and a one-shot example as the prompt. GPT-2 and Flan-T5 were fine-tuned on the full SLURP training data as a reference.}
    \centering
    \begin{tabular}{lccccc}
    \toprule
    Whisper Model     & Tiny (\%) & Base (\%) & Small (\%) & Medium (\%) & Reference (\%) \\
    WER & 33.4 & 26.4 & 19.8 & 14.6 & 0.0\\
    \midrule
    GPT-3.5  & 34.7 (32.0/38.0) & 36.7 (34.1/39.7) & 38.6 (35.0/42.7) & 40.8 (36.3/46.7) & 46.5 (40.5/54.5) \\
    GPT-4  & 41.3 (43.5/39.3) & 42.3 (42.1/42.5) & 45.2 (43.1/47.6)& 47.0 (43.7/50.9) & 53.6 (49.4/58.6) \\
    Vicuna-13B & 7.6 (4.5/20.1) & 7.7 (5.0/17.0) & 8.1 (5.2/17.9) & 8.9 (5.8/19.3) & 16.5 (11.5/29.5)\\
    Vicuna-13B-v1.5 & 16.6 (14.9/18.8) & 17.8 (16.3/19.6) & 20.7 (18.5/23.6) & 19.9 (17.3/23.3) & 22.4 (18.4/28.5)\\
    \midrule
    GPT-2 Full data & 54.9 (66.0/47.0) & 58.4 (70.6/49.9) & 64.0 (72.6/57.3) & 65.2 (74.1/58.2) & 84.8 (85.5/84.1) \\
    Flan-T5-base Full data & 56.4 (69.2/47.6) & 59.1 (72.2/50.8) & 64.6 (74.3/57.2) & 66.5 (76.7/58.8) & 86.3 (87.1/85.5)\\
    \bottomrule
    \end{tabular}
    \label{tab:asrerror}
\end{table*}

\subsection{Training and Inference Configurations}
Five LLMs were finetuned using LoRA with a rank of 8 in limited data setups, and with a rank of 128 when finetuning on the full dataset to give a similar number of trainable parameters as the baselines. During training, the LKI part was organised by selecting entities that appeared in the reference transcription, and adding random distractors found in the ASR hypotheses of the same utterance. All systems were trained on a single A100 GPU which took 2 hours to finetune LLaMA-13B under the 2000 sample limited data setup. 

During inference, a greedy search algorithm was used for all LLMs until the end-of-sequence token was output. The result was then fed to a post-processing stage using regular expressions to extract slot-value pairs in JSON format from the output text. As LLMs may fail to follow the exact instructions, for any utterances that fail to extract valid JSON format text, an empty output would be given. The LKI part was extracted based on the $N$-best hypotheses from the corresponding Whisper model. Systems were evaluated using the SLU-F1 metric, which combines both word-level and character-level F1 scores to give partial credit to non-exact match predictions.

\section{Results and Discussion}
\label{ssec:results}

\subsection{In-context learning with noisy transcriptions}

The performance of LLMs on slot filling via in-context learning with one-shot prompts is summarised in Table \ref{tab:asrerror} using different Whisper models. Note that LLaMA-13B without fine-tuning was not able to follow instructions and hence not able to perform slot filling. In general, LLMs yielded a higher recall rate than precision as they were not trained with the confined scenario setting and tended to pick up all possible fillers for each slot based on their knowledge. This problem became more severe when the definition of the slot type was rather abstract, such as \texttt{event\_name} or \texttt{news\_topic}, where LLMs tended to extract almost the entire utterance as the filler. The problem was less severe for slots requiring a single named entity to fill, e.g. \texttt{person}. For all systems, the degradation in recall was more than the degradation in precision when the ASR error rate increased, as important entities tended to be incorrectly recognised by the generic Whisper systems. Note that the degradation due to noisy transcriptions would probably be much smaller with an ASR system finetuned on SLURP audio with only a slightly lower overall WER \cite{KA2G}, as it is more likely to recognise domain-specific entities correctly.

Compared to the Vicuna-13B models containing 13B model parameters, GPT-3.5 and GPT-4 with much larger model sizes achieved much better results without any further parameter updates, which demonstrated the emergent ability of LLMs when the model size reached a certain level. GPT-4 achieved around a 7\% increase in SLU-F1 compared to GPT-3.5 under different ASR conditions while yielding a better balance between precision and recall. While GPT-2 and Flan-T5-base with task-specific fine-tuning achieved much better results with reference transcriptions, the performance degradation of using noisy transcription is also much larger compared to the non-finetuned LLMs. 

\begin{table}[h]
\caption{SLU-F1 scores of LLMs using few-shot in-context learning on SLURP test set with the reference or Whisper medium model transcriptions. GPT-2 and Flan-T5 models were fine-tuned on the number of samples indicated. Each sample took 20 tokens on average. Vicuna-13B was unable to have 100 examples in the prompt due to the model's maximum context lengths.}
    \centering
    \footnotesize
    \begin{tabular}{lccc}
    \toprule
    Systems     & N-samples & Medium (\%) & Ref. (\%) \\
    \midrule
    GPT-3.5  & 0 & 40.1 & 45.9 \\
    GPT-4  & 0 & 46.4 & 53.0 \\
    Vicuna-13B & 0 & 7.6 & 12.2\\
    Vicuna-13B-v1.5 & 0 & 17.6 & 18.3 \\
    \midrule
    GPT-3.5  & 1 & 40.8 & 46.5 \\
    GPT-4  & 1 & 47.0 & 53.6 \\
    Vicuna-13B & 1 & 8.9 & 16.5 \\
    Vicuna-13B-v1.5 & 1 & 19.9 & 22.4 \\
    \midrule
    GPT-3.5  & 10 & 42.3 & 46.8 \\
    GPT-4  & 10 & 49.9 & 55.9 \\
    Vicuna-13B & 10 & 11.8 & 21.4 \\
    Vicuna-13B-v1.5 & 10 & 21.3 & 24.9 \\
    \midrule
    GPT-3.5  & 100 & 47.4 & 53.5 \\
    GPT-4  & 100 & \textbf{57.4} & {65.8}\\
    Vicuna-13B-v1.5 & 100 & 25.8 & 31.3\\
    \midrule
    GPT-2* & 100 & 33.5 & 38.9 \\
    GPT-2* & 2000 & 52.2 & 65.2 \\
    Flan-T5-base* & 2000 & 55.2 & \textbf{68.9} \\
    \bottomrule
    \end{tabular}
    \label{tab:incontext}
\end{table}

An important aspect of in-context learning is to offer examples for LLMs to learn from them. To investigate the effect of the number of in-context learning samples, the performance against the number of samples for different systems is shown in Table \ref{tab:incontext}. 
As the number of samples in the prompt increased (except for the 100-sample case with Vicuna-13B which exceeded the maximum token lengths of the model), there was an increase in the SLU-F1 found in all LLMs both with reference and noisy transcriptions, at a cost of increasing prompt sequence lengths. In particular, GPT-4 achieved an obviously larger improvement than GPT-3.5 when more samples were included, indicating its superior capability in leveraging context information in the prompt. Notably, GPT-4 with 100 samples in-context learning performed much better than a GPT-2 model fine-tuned on the same number of samples and also surpassed the performance of a Flan-T5-base model fine-tuned on 2000 samples when using noisy transcription derived from the Whisper medium model. However, with 100 samples, the length of the prompt with 100 samples is close to 3000 in contrast to the 10-sample prompt which is around 1000. This significantly increased the inference cost with API calls.

\subsection{Task-specific fine-tuning}

\begin{figure}[h]
    \centering
    \hspace{-0.3cm}
    \includegraphics[scale=0.35]{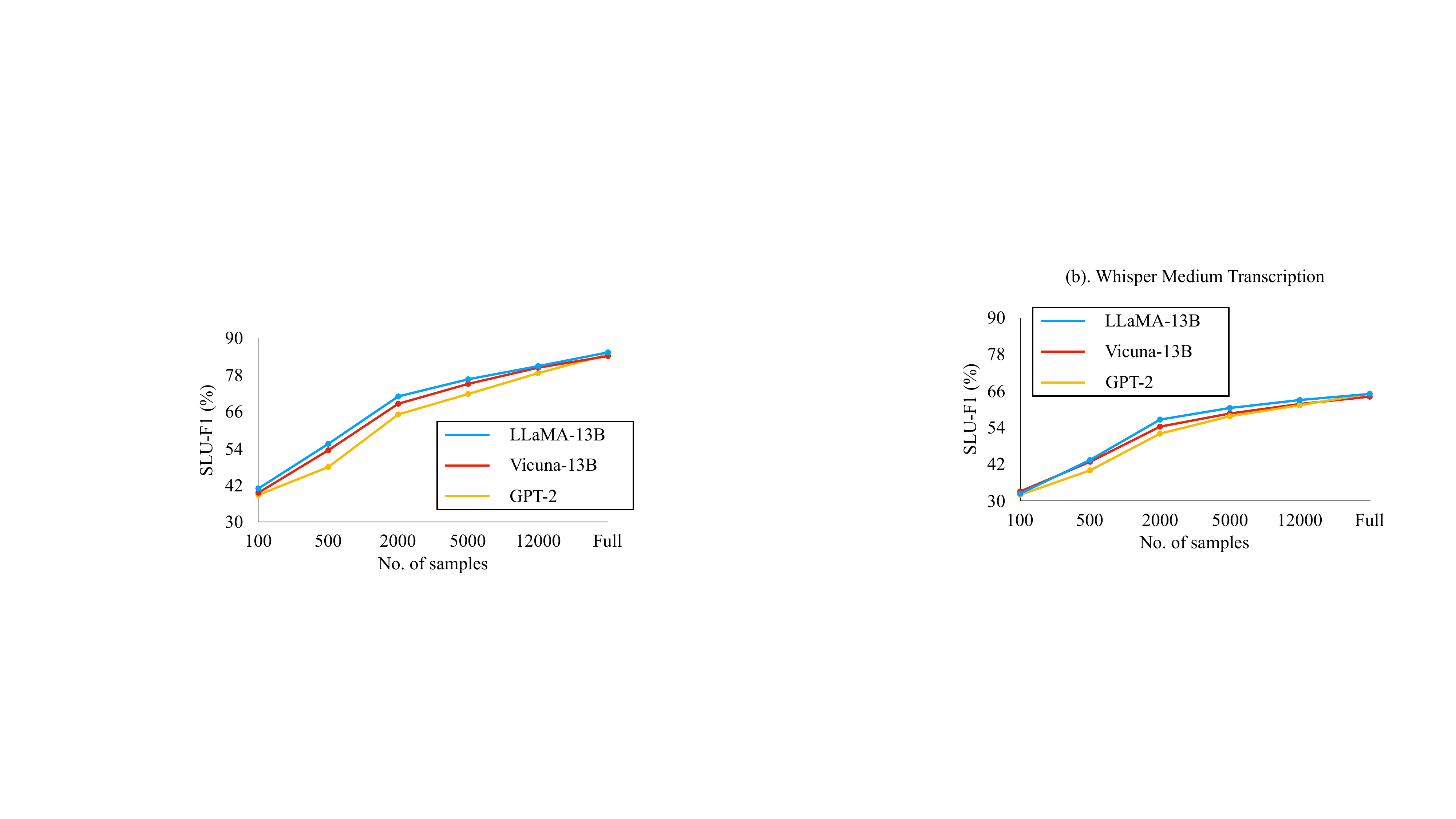}
    \caption{Variation of SLU-F1 on SLURP test set under reference transcription against the number of samples in the training set for fine-tuning three example LLMs.}
    \label{fig:trainset}
\end{figure}

Next, task-specific fine-tuning was applied to LLaMA-13B and Vicuna-13B v1.1 using LoRA on subsets of the SLURP training set for limited data scenarios. To begin with, SLU-F1 on the SLURP test set against different training set sizes was plotted as shown in Fig. \ref{fig:trainset}. With 500 and 2000 training samples representing limited data scenarios (2\% and 8\% of the full training set respectively), the performance of LLaMA-13B and Vicuna-13B achieved much better performance than GPT-2. The performance of Vicuna-13B was slightly worse than LLaMA-13B. When more samples were included, LLMs fine-tuned with LoRA gradually lost their advantage over the fully fine-tuned LMs. Further improvements may require full model fine-tuning for LLMs which necessitates distributed model parameters across multiple GPUs. In the following experiments, the 2000-sample limited data setup was used where LLMs achieved better performance than the best in-context learning performance achieved by GPT-4. A similar trend was observed under Whisper medium model transcriptions (see Appendix \ref{sec:appendix}).

\begin{table}[t]
\caption{SLU-F1 scores on the official SLURP test set and the zero-shot test set using LLaMA-13B, Flan-T5-base and GPT-2 fine-tuned on 2000 training samples. Reference transcriptions were used.}
\vspace{0.1cm}
    \centering
    \footnotesize
    \begin{tabular}{lccc}
    \toprule
    Systems     & Task desc. & Std. Test & Zero-shot Test \\
    \midrule
    GPT-2 & $\newcrossmark$ & 65.2\% & 0.0\% \\
    GPT-2 & $\newcheckmark$ & 65.7\% & 8.9\% \\
    Flan-T5-base & $\newcrossmark$ & 68.9\% & 0.0\% \\
    Flan-T5-base & $\newcheckmark$ & 69.8\% & 4.3\% \\
    LLaMA-13B & $\newcrossmark$ & 68.2\% & 4.9\% \\
    LLaMA-13B & $\newcheckmark$ & \textbf{71.1\%} & \textbf{52.1\%} \\
    \bottomrule
    \end{tabular}
    \label{tab:schema}
\end{table}

Then, \textbf{the effect of the task description} is demonstrated in Table \ref{tab:schema} for LLaMA-13B, Flan-T5-base and GPT-2 fine-tuned with 2000 training samples. Note that the training samples for the zero-shot setup did not contain any held-out slot types. While the task description had a limited influence on GPT-2 and a slightly larger influence on the Flan-T5 model, it achieved 3\% absolute SLU-F1 improvement in LLaMA-13B as LLMs were better at leveraging the context. Moreover, the importance of using a task description in the prompt in fine-tuning was also reflected in the zero-shot setup. By adding unseen slot descriptions to the prompt, the LLaMA-13B model was able to achieve an SLU-F1 score of 52.1\%, in contrast to the 4.9\% SLU-F1 from LLaMA-13B without the task description. The task description also helped the GPT-2 and Flan-T5-base models on unseen slot types, but the improvement was very limited and the performance was far behind that of the LLaMA-13B model. This showcases the emergent ability of LLMs reflected in zero-shot learning.

\begin{table}[t]
\caption{SLU-F1 (Precision/Recall) on SLURP test set using transcriptions from the Whisper medium model and linearised knowledge injection (LKI) in both in-context learning and fine-tuning (indicated with FT) setups. Fine-tuning was performed on the 2000-sample subset. Note that 13B was omitted for clarity in this table.}
\vspace{-0.1cm}
    \centering
    \footnotesize
    \begin{tabular}{lcc}
    \toprule
    Systems     & With LKI (\%) & No LKI (\%) \\
    \midrule
    GPT-3.5 & 46.8 (36.1/66.4) & 40.8 (36.3/46.7) \\
    Vicuna & 21.8 (15.3/37.4) & 8.9 (5.8/19.3) \\
    Vicuna-v1.5 & 33.0 (27.3/41.8) & 19.9 (17.3/23.3) \\
    \midrule
    GPT-2 FT & 57.0 (65.8/50.2) & 52.2 (61.8/45.0) \\
    Flan-T5-base FT & 59.5 (68.3/52.7) & 55.2 (66.3/47.3) \\
    LLaMA FT & \textbf{62.6} (71.2/55.9) & \textbf{56.7} (69.2/48.0) \\
    Vicuna FT & 61.3 (70.0/54.5) & 53.4 (61.5/47.2) \\
    Vicuna-v1.5 FT & 62.5 (72.5/55.0) & 55.0 (65.5/47.3) \\
    \bottomrule
    \end{tabular}
    \vspace{-0.2cm}
    \label{tab:lki}
\end{table}

\textbf{Linearised contextual knowledge} provided important external information about the application context. To this end, the effect of such knowledge injected using the proposed LKI approach was investigated as shown in Table \ref{tab:lki} for both in-context learning and fine-tuning. As shown in Table \ref{tab:lki}, the main effect of LKI was to improve the recall as it mainly provided information for entities that were left out by the system. LKI for GPT-3.5 had a rather limited effect, as while improving the recall, it decreased the precision as noise in the LKI prompt introduced many irrelevant slot values. The influence of LKI on Vicuna-13B under in-context learning almost doubled the SLU-F1, although the numbers were still much lower than GPT-3.5. When fine-tuned with LKI in the prompt, all LMs achieved larger improvements in the overall SLU-F1, as fine-tuned models had a much lower recall. The Vicuna-13B-v1.5 benefited the most from LKI as it is able to process long context better, whereas the best performance was still achieved by LLaMA-13B. As a result, the proposed LKI approach proved to be an effective way for dynamic contextual knowledge integration in LLMs.

\subsection{Fine-tuning with noisy transcriptions}

\begin{table}[t]
\caption{SLU-F1 on SLURP test set with LLMs fine-tuned on noisy transcriptions from Whisper medium model. The training settings indicate the number of hypotheses concatenated and used in the prompt. The same number of hypotheses in training was used during inference. All LLMs were fine-tuned and fine-tuning was performed on the 2000-sample subset.}
    \centering
    \footnotesize
    \begin{tabular}{lcc}
    \toprule
    System & Training & SLU-F1 (\%) \\
    \midrule
    GPT-2 & ref & 52.2 (61.8/45.0) \\
    GPT-2 & 10-best & 39.0 (62.3/28.3)\\
    Flan-T5-base & ref & 55.2 (66.3/47.3) \\
    Flan-T5-base & 10-best & 35.2 (59.2/25.1)\\
    \midrule
    LLaMA-13B & ref. & 56.7 (69.2/48.0)\\
    LLaMA-13B & 1-best & 54.5 (66.5/46.1)\\
    LLaMA-13B & 3-best & 57.5 (63.3/52.7)\\
    LLaMA-13B & 5-best & 58.1 (62.9/54.0)\\
    LLaMA-13B & 10-best & \textbf{60.6} (65.3/56.6)\\
    LLaMA-13B & 20-best & 60.0 (65.7/55.2)\\
    LLaMA-13B & 9-best + ref. & 59.0 (64.3/54.5) \\
    \midrule
    LLaMA-13B LKI & 10-best & \textbf{63.5} (67.9/57.6) \\
    \bottomrule
    \end{tabular}
    \label{tab:nbest}
\end{table}


The final experiments were performed in order to explore the use of multiple hypotheses from the ASR system to improve the robustness of LLMs to ASR errors for slot filling. The results of using different numbers of hypotheses are summarised in Table \ref{tab:nbest}. As shown in the first part of Table \ref{tab:nbest}, the best-performing number of hypotheses was 10, which achieved a 4\% absolute SLU-F1 increase compared to the single-hypothesis case, with more balanced precision and recall. However, using the 10-best hypotheses in GPT-2 and Flan-T5-base confused the model and yielded much worse results with a severely degraded recall rate.

Ablation studies were performed as shown in the second part of Table \ref{tab:nbest}. Although using references for training achieved better performance than the 1-best hypotheses, incorporating the references into $N$-best lists for training yielded worse performance as the model still relied on the reference presented in the $N$-best list. 

The best SLU-F1 score was achieved by combining the proposed LKI together with the proposed fine-tuning using noisy transcriptions, with an overall \textbf{6.8}\% absolute SLU-F1 increase compared to using the 1-best prompt and \textbf{8.3}\% absolute increase compared to the strong Flan-T5-base baseline system. Although the knowledge was selected based on the $N$-best hypotheses, LKI was still informative when the $N$-best list was included in the context and achieved a 3\% absolute increase in SLU-F1. 

\section{Conclusions}
\label{sec:conclusion}
This paper investigated the performance of LLMs for slot filling with noisy transcriptions from speech. It quantified the performance of five different LLMs using ASR transcriptions from four different sizes of Whisper ASR models and proposed a dedicated prompt design, a noise-robust fine-tuning approach and a linearised knowledge integration (LKI) scheme. In particular, the proposed prompt design with few-shot examples enabled GPT-4 to outperform a GPT-2 model fine-tuned on 20 times more data. Meanwhile, the LLaMA-13B model achieved an absolute 8.3\% SLU-F1 increase using both the noise-robust fine-tuning and the LKI scheme compared to a strong fully fine-tuned Flan-T5-base model.

\section{Limitation}
\label{sec:limit}

The main limitation of using LKI and n-best hypotheses for training is the context length of LLMs. When sequence length became much longer by incorporating more examples or more hypotheses, the efficacy of long context is inherently limited by the effective spans of the attention mechanisms in LLMs. Even with the Vicuna-13B-v1.5 model which supports a sequence length of 16k tokens, a clear reduction in improvements was found when adding much more information into the context. Therefore, future work needs to be done on improving the effectiveness of the context instead of only increasing the allowed maximum sequence length.

\section{Ethics Statement}
The approaches in this paper do not give rise to any additional risks beyond the ones directly inherited from the models. The ASR system might work worse for speakers from particular demographics and of particular accents. The framework also inherits the biases from all the language models used for experiments in this paper.

\bibliography{anthology,custom}
\bibliographystyle{acl_natbib}

\newpage
\appendix

\section{Training set sizes}
\label{sec:appendix}

\begin{figure}[h]
    \centering
    \hspace{-0.3cm}
    \includegraphics[scale=0.35]{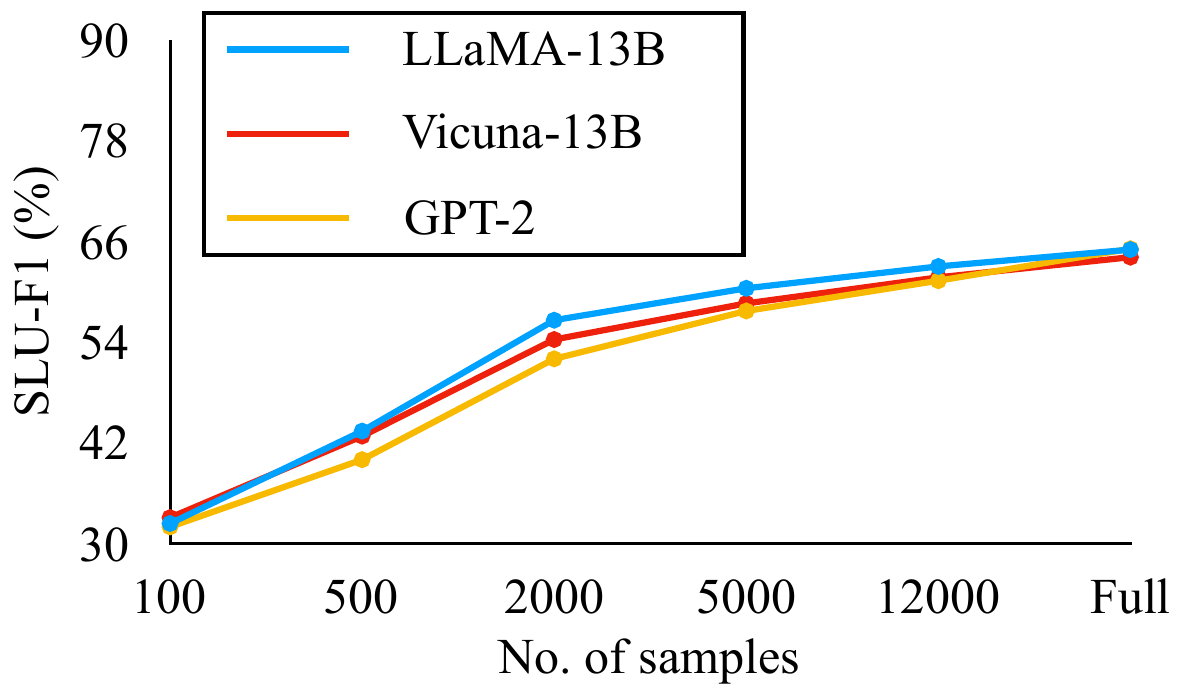}
    \caption{Variation of SLU-F1 on SLURP test set under Whisper medium model transcription against the number of samples in the training set for fine-tuning LLMs.}
    \label{fig:trainset2}
\end{figure}

The influence of training set sizes on different LLMs and GPT-2 under Whisper medium ASR transcription is shown in Fig. \ref{fig:trainset2}.

\end{document}